\documentclass[journal]{IEEEtran}

\usepackage{amsmath,amsfonts,amssymb}
\usepackage{array}
\usepackage{booktabs}
\usepackage{cite}
\usepackage{graphicx}
\usepackage{float}
\usepackage{multirow}
\usepackage{url}

\title{Geometry-Preserving Human-to-Robot Upper-Body Motion Retargeting from Monocular Video}

\author{
Xiaoyu Yang,
Sen Han,
Da Li,
and Nan Wu
\thanks{
Xiaoyu Yang and Nan Wu are with
Across Physics,
Beijing, China.
}
\thanks{
Sen Han and Da Li are with
Beijing SFlare Robotics Technology Co., Ltd.,
Beijing, China.
}
\thanks{
Corresponding author: Nan Wu
(e-mail: across2026@163.com).
}
}

\begin{document}

\maketitle

\begin{abstract}
Monocular RGB video provides an accessible source of human
demonstrations for upper-body robot motion, yet video-driven
human-to-robot transfer remains challenging because body and hand
motion are recovered at different spatial scales, human and robot
kinematics differ substantially, and fine distal motion is difficult
to preserve across embodiments. We present a geometry-preserving
motion-retargeting framework that integrates unified body--hand
reconstruction with morphology-independent geometric transfer.
Frame-wise body estimates, video-level observations, and detailed hand
evidence jointly constrain a single differentiable Momentum Human Rig
(MHR) state, while transient hand artifacts are repaired in parameter
space. The reconstructed motion is represented by arm-segment
directions, elbow configuration, relative palm orientation, and
bilateral wrist relations, and is realized on the target robot through
multi-stage inverse kinematics and robot-specific hand adaptation.
Within the broader system, Across-VAM provides video generation,
whereas Across-WAM performs human-to-robot motion mapping. The method
is evaluated on 16 monocular videos comprising 1,769 source frames,
including 10 signing and six reach-to-grasp sequences. Unified
reconstruction reduces mean hand reprojection error from 22.36 to
7.21 pixels relative to SAM 3D Body. All 16 retargeted trajectories
completed kinematic simulation playback, and representative signing
and reach-to-grasp motions were further demonstrated on a physical
robot. The results demonstrate a unified pipeline from monocular human
video to coordinated upper-body motion on a dual-arm dexterous robot.
\end{abstract}


\section{Introduction}
\label{sec:introduction}

Human motion provides a rich source of demonstrations for robots.
Acquiring such demonstrations commonly relies on motion-capture
systems, wearable devices, teleoperation interfaces, or task-specific
instrumentation~\cite{qin2023anyteleop,wang2024dexcap}. Monocular RGB
video provides a widely accessible alternative for capturing natural
human behavior. Recovering motion from such videos and transferring it
to robots with different kinematics is therefore an important problem
in robot programming and human--robot interaction.

This capability is particularly relevant to assistive applications
that involve both communicative and physical behaviors. Sign-language
reproduction requires coordinated bilateral arm motion, palm
orientation, and distinctive hand configurations, whereas
reach-to-grasp behavior relies on coordinated motion of the arm,
wrist, and hand toward an interaction target~\cite{gago2019teo,
huo2025srl}. Prior work has studied latent optimization for sign
retargeting~\cite{zhang2022kinematic}, video-driven sign transfer
~\cite{zhang2024video}, humanoid sign reproduction
~\cite{gago2019teo}, dedicated anthropomorphic signing hands
~\cite{bulgarelli2016hand}, and tactile fingerspelling systems
~\cite{johnson2021tatum}. Human demonstrations have also supported
functional multifinger grasping~\cite{wei2024functional}, reactive
dexterous handover~\cite{duan2025handover}, and long-horizon dual-arm
manipulation~\cite{kim2024goal}. Although signing and reaching stress
different aspects of motion, both require task-relevant geometry to be
preserved across different link lengths, joint arrangements, and hand
actuation mechanisms.

A central challenge lies in obtaining a coherent human representation
from monocular video. Global body motion and fine hand articulation
appear at substantially different spatial scales and exhibit different
estimation characteristics. HMR established direct regression of
parametric human shape and pose from RGB images
~\cite{kanazawa2018hmr}, and HMR~2.0 extended transformer-based mesh
recovery to video tracking~\cite{goel2023humans}. Component-aware
whole-body reconstruction allocates representation capacity across
body regions~\cite{lin2023osx}, while Hand4Whole emphasizes dedicated
features for fine hand articulation~\cite{moon2022hand4whole}.
Hand-specific models provide complementary distal detail. MANO offers
a compact parametric hand model~\cite{romero2017mano}, and HaMeR,
Deformer, and Dyn-HaMR advance monocular hand reconstruction, temporal
fusion, and interacting-hand recovery~\cite{pavlakos2024hamer,
fu2023deformer,yu2025dynhamr}.

Recent full-body representations provide a common state for combining
these observations. SAM 3D Body reconstructs body and hands with the
Momentum Human Rig (MHR)~\cite{yang2026sam3dbody,ferguson2025mhr},
while SAM-Body4D incorporates video-level temporal information
~\cite{gao2025sambody4d}. For video-to-robot transfer, detailed hand
evidence should remain kinematically coupled with the wrist, forearm,
elbow, and shoulder rather than being attached to the body after
independent estimation.

Cross-embodiment realization introduces a second challenge. The same
human motion can correspond to substantially different robot joint
configurations because of morphological differences. Consequently,
the choice of preserved geometric attributes becomes central to
retargeting. Optimization-based whole-body methods use task-space
quantities and robot kinematics~\cite{penco2018retargeting}.
DexPilot and AnyTeleop extend this principle to coordinated arm--hand
control~\cite{handa2020dexpilot,qin2023anyteleop}, while DexCap and
LEGATO introduce embodiment-aware representations for demonstration
transfer and acquisition~\cite{wang2024dexcap,seo2025legato}. General
Motion Retargeting studies morphology-aware references for whole-body
tracking~\cite{araujo2026gmr}, and BeyondRetarget explores direct
monocular-video-to-humanoid generation with shared visual features and
robot-specific decoders~\cite{xiong2026beyondretarget}.
Spatio-temporal retargeting further separates morphology-aware spatial
adaptation from temporal refinement~\cite{yoon2025spatiotemporal}.
These results motivate an explicit geometric representation whose
spatial realization and temporal conditioning can be evaluated
separately.

We address these challenges through a two-stage framework for
geometry-preserving upper-body motion retargeting. First, body-level
and hand-level observations jointly optimize a single differentiable
Momentum Human Rig (MHR) state. This unified representation couples
fine hand reconstruction with the complete supporting upper-body
kinematic chain. Transient distal artifacts are subsequently corrected
in MHR parameter space while preserving the articulated representation.

Second, the reconstructed human motion is expressed through
morphology-independent geometric attributes, including upper-arm and
forearm directions, elbow configuration, palm orientation, and
bilateral wrist relations. Multi-stage numerical IK reconstructs these
attributes through the target robot's forward kinematics. Each frame
is solved from a common initialization, followed by command-space
temporal conditioning. Distal articulation is adapted through
robot-specific handshape presets that match the coupled actuation
space of the dexterous hands.

We evaluate the framework on 16 monocular videos comprising 10
signing and six reach-to-grasp sequences. Signing
emphasizes bilateral coordination, palm orientation, and distinctive
hand configurations, while reach-to-grasp emphasizes coordinated
arm--wrist--hand motion during object-directed behavior. Together,
the two categories evaluate geometric retargeting from monocular
human video to a dual-arm robot.

The main contributions of this work are threefold:

\begin{itemize}
    \item We formulate monocular upper-body reconstruction as a
    unified optimization over a single articulated MHR state, in
    which global body observations and detailed hand observations
    jointly constrain the connected body--hand kinematic chain.
    Model-space repair further improves the temporal consistency of
    distal articulation.

    \item We introduce a morphology-independent retargeting
    formulation that preserves arm-segment directions, elbow
    configuration, relative palm orientation, and bilateral wrist
    relations. Multi-stage numerical IK reconstructs these attributes
    through the target robot's own kinematics.

    \item We systematically evaluate the framework on 16 monocular
    sequences covering signing and reach-to-grasp motion.
    Sequence-level experiments demonstrate improvements in human
    reconstruction, cross-embodiment geometric transfer, and
    temporal command regularity.
\end{itemize}

\section{Method Overview}
\label{sec:overview}

\begin{figure*}[t]
\centering
\includegraphics[width=0.784\textwidth]{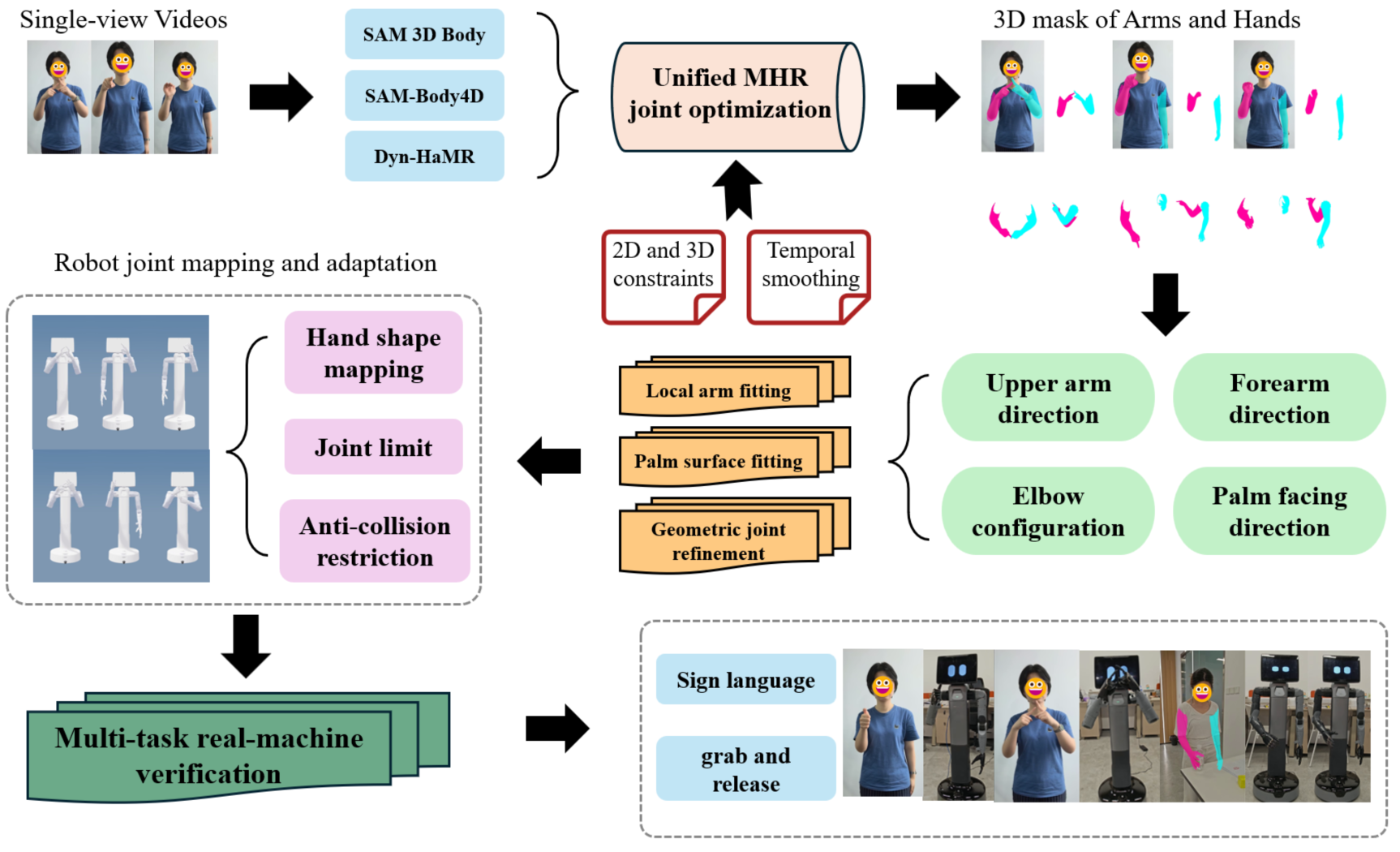}
\caption{Overall framework for retargeting motion from monocular human
video to a dual-arm dexterous robot.}
\label{fig:figure_3}
\end{figure*}

The proposed framework addresses upper-body motion retargeting from
monocular human video to a dual-arm dexterous robot. It comprises human
motion reconstruction, cross-embodiment geometric mapping, robot-joint
adaptation, and task validation, as shown in
Fig.~\ref{fig:figure_3}. The input is a monocular video acquired from a
fixed viewpoint. SAM 3D Body, SAM-Body4D, and Dyn-HaMR respectively
provide frame-wise human states, video-level body observations, and
detailed hand observations. These signals jointly constrain the same
differentiable MHR model. The optimization incorporates 2-D
reprojection, local 3-D hand geometry, and temporal consistency to
produce a coherent 3-D upper-limb--hand state for retargeting.

The unified human state is converted into cross-embodiment geometric
attributes. They include upper-arm and forearm directions, elbow
configuration, palm orientation, and, for bilateral actions, the
relative relation between the two wrists. These quantities adapt more
naturally to differences in scale and link length than absolute human
joint positions. The target robot uses its own kinematic model to fit
local arm geometry, palm orientation, and the combined geometric
objectives in successive stages.

Robot-specific adaptation follows geometric mapping. Handshape
information is extracted from the reconstructed 3-D hands and mapped
to calibrated dexterous-hand commands. Arm and hand outputs obey robot
joint limits, and bilateral poses are adjusted when necessary to
maintain geometric clearance. Temporal conditioning then suppresses
high-frequency changes introduced by frame-wise numerical optimization.
The same framework handles signing and reach-to-grasp motion. The
former emphasizes bilateral coordination, palm changes, and discrete
handshapes, whereas the latter emphasizes unilateral approach, wrist
adjustment, and grasp-state transitions. Generated motions are first
validated kinematically in simulation and are then demonstrated on a
physical robot.

\section{Body--Hand Reconstruction}
\label{sec:reconstruction}

The reconstruction stage estimates a single articulated human state
that jointly explains global upper-body structure and local hand
geometry. SAM 3D Body provides frame-wise MHR initialization,
SAM-Body4D contributes video-level body observations, and Dyn-HaMR
provides confidence-weighted 2-D hand landmarks together with
wrist-local 3-D hand geometry.

SAM-Body4D contributes image-space MHR70 observations and frame
validity to the joint optimization. Human shape, scale, expression,
and focal length are robustly aggregated over valid frames and fixed
throughout sequence optimization. The optimized variables comprise 20
bilateral clavicle-to-wrist body parameters, two 54-dimensional
hand-pose blocks, and per-frame camera-translation increments, while
the remaining body-pose parameters and global body orientation stay at
their frame-wise SAM 3D Body initialization.

Let $\mathbf J^b_{t,j}$ be body landmark $j$ produced by the MHR
forward model at frame $t$, and let $\hat{\mathbf u}^b_{t,j}$ be its
image-space observation. Define $\mathbf J^h_{t,i}$ and
$\hat{\mathbf u}^h_{t,i}$ analogously for hand landmarks. With camera
parameters $\gamma_t$, the body and hand reprojection terms are
\begin{equation}
\begin{aligned}
\mathcal L_{\mathrm{body}}&=\sum_{t,j}c^b_{t,j}\rho\!\left(
\left\|\Pi_{\gamma_t}(\mathbf J^b_{t,j})-
\hat{\mathbf u}^b_{t,j}\right\|_2^2\right),\\
\mathcal L_{\mathrm{hand}}^{2D}&=\sum_{t,h,i}c^h_{t,i}\rho\!\left(
\left\|\Pi_{\gamma_t}(\mathbf J^h_{t,i})-
\hat{\mathbf u}^h_{t,i}\right\|_2^2\right).
\end{aligned}
\label{eq:reprojection}
\end{equation}
Here $h\in\{L,R\}$, $c^b_{t,j}$ and $c^h_{t,i}$ are observation
confidences, $\Pi_{\gamma_t}$ is perspective projection, and $\rho$ is
a robust loss. Body observations from SAM 3D Body and SAM-Body4D enter
$\mathcal L_{\mathrm{body}}$, while all 21 Dyn-HaMR landmarks form the
fine-grained hand term.

Local 3-D hand observations complement image-space reprojection by
providing constraints on finger flexion, palm structure, and relative
depth. We use wrist-centered observations normalized by palm scale.
Landmark 0 is the wrist,
and $\mathcal M=\{5,9,13,17\}$ contains the four metacarpophalangeal
joints. The scale, normalized landmarks, and 3-D loss are
\begin{equation}
\begin{aligned}
s_t^h&=\frac{1}{4}\sum_{i\in\mathcal M}
\left\|\mathbf J^h_{t,i}-\mathbf J^h_{t,0}\right\|_2,\\
\bar{\mathbf J}^h_{t,i}&=\frac{\mathbf J^h_{t,i}-
\mathbf J^h_{t,0}}{s_t^h},\\
\mathcal L_{\mathrm{hand}}^{3D}&=\sum_{t,h,i}c^h_{t,i}\rho\!\left(
\left\|\bar{\mathbf J}^h_{t,i}-
\hat{\bar{\mathbf X}}^h_{t,i}\right\|_2^2\right).
\end{aligned}
\label{eq:localhand}
\end{equation}
The observation $\hat{\bar{\mathbf X}}^h_{t,i}$ is normalized in the
same manner. This representation removes absolute translation and
body-scale effects while retaining finger configuration and palm
orientation. Because the normalized point cloud retains its observed
orientation, it also provides directional evidence to the wrist and
adjacent upper-limb chain.

Priors on pose and camera-translation increments prevent excessive
deviation from initialization. Second-order temporal differences are
applied to arm landmarks, hand landmarks, and camera translation. For
any trajectory $\mathbf P_t$, the difference is
$\mathbf P_{t+1}-2\mathbf P_t+\mathbf P_{t-1}$. The complete objective
is
\begin{equation}
\begin{aligned}
\mathcal{L}_{\mathrm{MHR}}
={}&
\lambda_b\mathcal{L}_{\mathrm{body}}
+
\lambda_{2D}\mathcal{L}_{\mathrm{hand}}^{2D}
+
\lambda_{3D}\mathcal{L}_{\mathrm{hand}}^{3D}\\
&+
\lambda_p\mathcal{L}_{\mathrm{prior}}
+
\lambda_t\mathcal{L}_{\mathrm{temp}} .
\end{aligned}
\label{eq:mhrloss}
\end{equation}
Optimization follows two stages. The first stage refines the bilateral
arm parameters and camera translation so that wrist observations
adjust the connected shoulder--elbow--wrist chain. The second stage
releases the hand-pose variables and incorporates the complete 21-point
2-D and local 3-D hand observations. Because all terms share the same
differentiable MHR state, distal evidence contributes directly to the
connected wrist, forearm, elbow, upper arm, and clavicle configuration.
The optimized sequence provides one coherent body--hand representation
for subsequent retargeting.

\subsection{Model-Space Hand Repair}

Monocular hand reconstruction occasionally produces a short-lived
departure from an otherwise coherent trajectory. We characterize this
pattern through temporal round-trip consistency: a candidate frame
exhibits a sufficiently large displacement from both neighboring
states while the two neighboring states remain mutually consistent.
Observation confidence, finite-depth validity, and motion-direction
consistency provide additional cues for identifying transient distal
artifacts.

For each detected frame, the corresponding MHR hand-pose parameter
block is reconstructed by interpolation between neighboring trusted
states. The complete mesh and landmark configuration are then
regenerated through the MHR forward model. Performing the correction
in parameter space preserves the articulated structure of the body and
hand while restoring a temporally coherent distal trajectory.

\subsection{Human Reconstruction Results}

\begin{figure*}[t]
\centering
\includegraphics[width=0.784\textwidth]{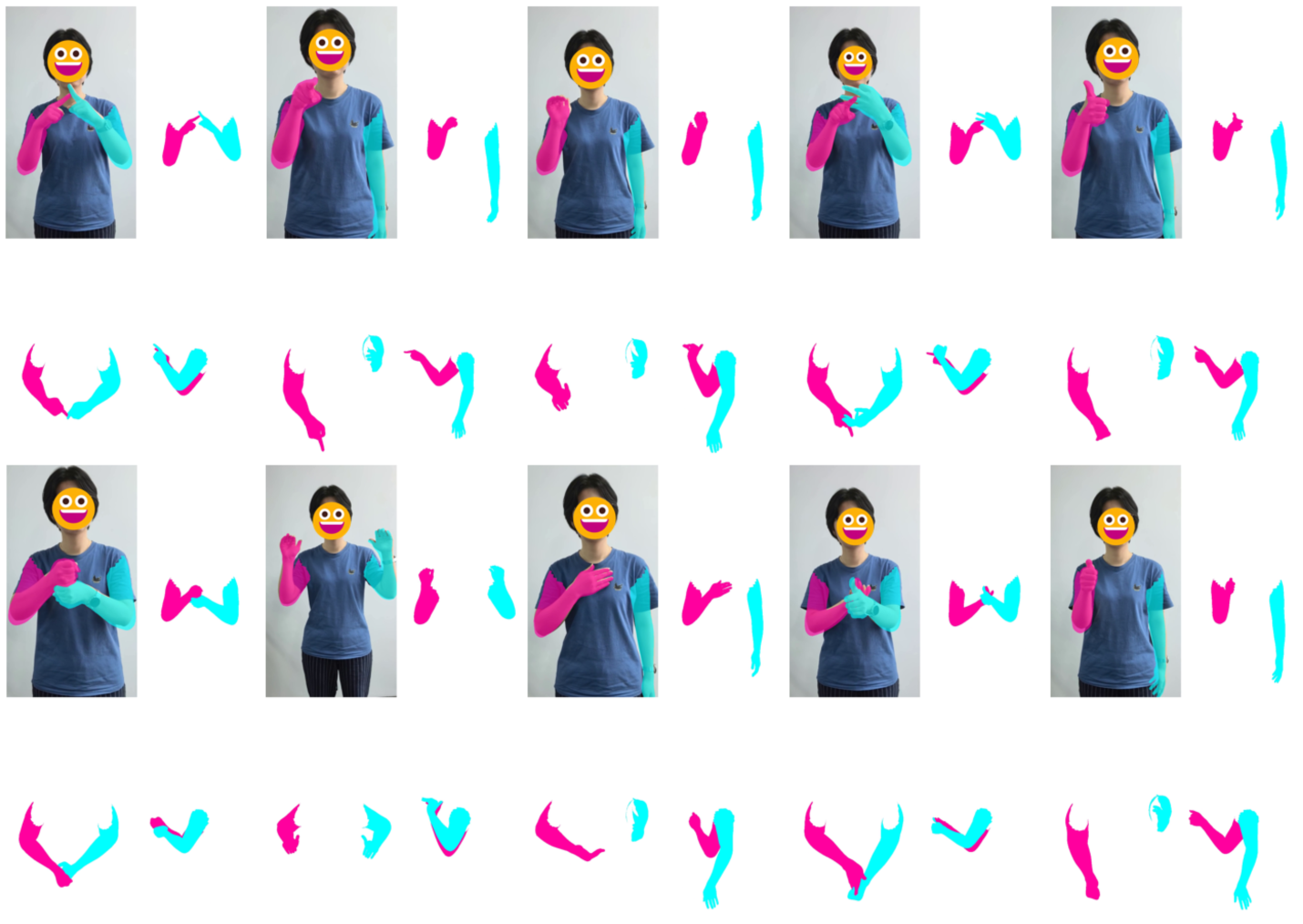}
\caption{Multi-view upper-limb--hand mask extraction for ten
representative signs. From left to right, the first row shows
\emph{person}, \emph{you}, \emph{can}, \emph{name}, and \emph{good};
the second row shows \emph{work}, \emph{meeting}, \emph{i},
\emph{love}, and \emph{thank you}. Each group contains the mask on the
source frame and additional views of the same reconstructed state.}
\label{fig:human_reconstruction}
\end{figure*}

Figure~\ref{fig:human_reconstruction} shows multi-view upper-limb and
hand mask extraction for ten representative signs. Each result is
generated by forwarding the same optimized MHR state. The body, arms,
and hands therefore share one articulated 3-D representation. The
examples cover unilateral and bilateral motion, varied elbow flexion,
wrist orientation, and multiple hand configurations. The connected
shoulder--elbow--wrist--hand state provides the geometric quantities
used by the subsequent retargeting stage.

\section{Cross-Embodiment Retargeting}
\label{sec:retargeting}

Cross-embodiment retargeting is formulated in terms of geometric
attributes that remain meaningful across different limb lengths,
joint placements, and actuation structures. The representation
contains local arm directions, elbow configuration, relative palm
orientation, and bilateral wrist relations. These attributes are
realized through the target robot's forward kinematics and numerical
inverse kinematics.

\subsection{Arm Geometry}

For side $h\in\{L,R\}$, let $\mathbf p^h_{S,t}$,
$\mathbf p^h_{E,t}$, and $\mathbf p^h_{W,t}$ denote the reconstructed
shoulder, elbow, and wrist locations in camera coordinates. A fixed
coordinate transform maps camera axes $(x,y,z)$ to robot-aligned
forward, lateral, and vertical axes $(-z,x,-y)$. Applying this common
transformation yields the upper-arm and forearm targets
\begin{equation}
\begin{aligned}
\mathbf u_t^{h,*} &= \mathbf R_{CR}
\frac{\mathbf p^h_{E,t}-\mathbf p^h_{S,t}}
{\|\mathbf p^h_{E,t}-\mathbf p^h_{S,t}\|_2},\\
\mathbf f_t^{h,*} &= \mathbf R_{CR}
\frac{\mathbf p^h_{W,t}-\mathbf p^h_{E,t}}
{\|\mathbf p^h_{W,t}-\mathbf p^h_{E,t}\|_2}.
\end{aligned}
\label{eq:armdirections}
\end{equation}
Here $\mathbf R_{CR}$ denotes the fixed camera-to-robot axis
conversion. The two segment directions further determine elbow
flexion $\theta_t^{h,*}$ and elbow-bending direction
$\mathbf b_t^{h,*}$:
\begin{equation}
\begin{aligned}
\theta_t^{h,*}&=\arccos\!\left[\mathrm{clip}\!\left(
(\mathbf u_t^{h,*})^\top\mathbf f_t^{h,*},-1,1\right)\right],\\
\mathbf b_t^{h,*}&=\frac{\mathbf f_t^{h,*}-
[(\mathbf u_t^{h,*})^\top\mathbf f_t^{h,*}]\mathbf u_t^{h,*}}
{\left\|\mathbf f_t^{h,*}-[(\mathbf u_t^{h,*})^\top
\mathbf f_t^{h,*}]\mathbf u_t^{h,*}\right\|_2}.
\end{aligned}
\label{eq:elbowgeometry}
\end{equation}
The bending direction receives an
observability-dependent weight that decreases as the arm approaches
full extension. Together, these quantities provide a translation- and
scale-invariant description of the source arm configuration. Robot
link lengths subsequently determine the corresponding spatial
realization through forward kinematics.

\subsection{Bilateral Relations}

Local arm geometry characterizes each arm, while bilateral motion also
depends on the relative configuration of the two hands. Let
$\mathbf s_t^L,\mathbf s_t^R$ and $\mathbf w_t^L,\mathbf w_t^R$ be the
2-D shoulder and wrist positions. With human shoulder width
$d_t^H=\|\mathbf s_t^L-\mathbf s_t^R\|_2$, the normalized horizontal
separation and signed height relation are
\begin{equation}
r_{x,t}^H=\frac{|w_{x,t}^R-w_{x,t}^L|}{d_t^H},\qquad
r_{y,t}^H=\frac{w_{y,t}^R-w_{y,t}^L}{d_t^H}.
\label{eq:bilateralrelation}
\end{equation}
Robot relations $r_{x,t}^R(\mathbf q)$ and $r_{y,t}^R(\mathbf q)$ are
normalized by robot shoulder width. After applying a common sign
convention, their loss is
\begin{equation}
\mathcal L_{\mathrm{bil},t}=\lambda_x(r_{x,t}^R-r_{x,t}^H)^2+
\lambda_y(r_{y,t}^R-r_{y,t}^H)^2.
\label{eq:bilateralloss}
\end{equation}
The height relation is bounded before optimization to prevent 2-D
scale variations from dominating the 3-D arm configuration. These
relational terms complement the arm-direction objectives and preserve
visually important cross-arm structure in two-handed motion.
For unilateral reach-to-grasp sequences, the active arm is solved
independently while the complementary arm retains its reference
configuration.

\subsection{Palm Orientation}

Each human palm frame uses MHR hand landmarks centered at the wrist.
The wrist and middle metacarpophalangeal (MCP) joint define the palm's
forward axis. The index and little MCP joints define its lateral
axis, and their cross product provides the palm normal. After
orthonormalization, the frame is expressed in
robot-aligned coordinates through $\mathbf R_{CR}$. Let
$\mathbf R_{p,t}^{H,h}$ denote the converted human palm frame and
$\mathbf R_{p,t_0}^{R,h}$ the robot palm frame at calibration frame
$t_0$. The target robot palm orientation is
\begin{equation}
\mathbf R_{p,t}^{h,*}
=\mathbf R_{p,t_0}^{R,h}
\bigl(\mathbf R_{p,t_0}^{H,h}\bigr)^{\top}
\mathbf R_{p,t}^{H,h}.
\label{eq:relativepalm}
\end{equation}
Equation~(\ref{eq:relativepalm}) transfers the palm rotation relative
to the calibration frame and accommodates the different mounting
orientations of the human and robot hands. The resulting target is
combined with arm geometry during numerical IK.

\subsection{Multi-Stage IK}

Robot forward kinematics provides upper-arm and forearm directions
$\mathbf u_t^{R,h}(\mathbf q)$ and $\mathbf f_t^{R,h}(\mathbf q)$,
elbow-bending direction $\mathbf b_t^{R,h}(\mathbf q)$, flexion
$\theta_t^{R,h}(\mathbf q)$, and palm orientation
$\mathbf R_{p,t}^{R,h}(\mathbf q)$ for a candidate joint
configuration. For the active arm set $\mathcal A_t$, the geometric
fitting objective is
\begin{equation}
\begin{aligned}
\mathcal L_t(\mathbf q)={}&\sum_{h\in\mathcal A_t}\Bigl[
\lambda_u\|\mathbf u_t^{R,h}-\mathbf u_t^{h,*}\|_2^2\\
&+\lambda_f\|\mathbf f_t^{R,h}-\mathbf f_t^{h,*}\|_2^2\\
&+w_{b,t}^h\lambda_b
\|\mathbf b_t^{R,h}-\mathbf b_t^{h,*}\|_2^2\\
&+\lambda_\theta(\theta_t^{R,h}-\theta_t^{h,*})^2\\
&+\lambda_R\bigl\|\log\bigl(
(\mathbf R_{p,t}^{h,*})^\top
\mathbf R_{p,t}^{R,h}\bigr)\bigr\|_F^2\Bigr]\\
&+\mathcal L_{\mathrm{bil},t}(\mathbf q).
\end{aligned}
\label{eq:geometryik}
\end{equation}
The bending-direction weight $w_{b,t}^h$ decreases near elbow
extension according to geometric observability. For bilateral motion,
$\mathcal L_{\mathrm{bil},t}$ contains the normalized inter-wrist
relations introduced in Section~V-B. Each stage respects the robot's
joint limits.

The numerical solution proceeds through four stages. First, each
active arm is fitted to its local direction and elbow targets. Second,
bilateral sequences undergo joint refinement using cross-arm wrist
relations. Third, palm orientation is fitted from the resulting arm
configuration. Finally, all active joints are jointly refined using
the complete geometric objective. Multiple initialization candidates
are evaluated at each frame, and the valid solution with the lowest
residual is retained. Each video frame uses the same robot reference
initialization. This frame-independent formulation provides
consistent numerical conditions throughout the sequence, after which
temporal regularity is introduced at the command level.

\subsection{Hand Adaptation and Temporal Conditioning}

The target dexterous hand exposes a coupled six-channel actuation
space. We represent distal articulation through a compact library of
robot-specific hand configurations. Finger-bending features are
extracted from the reconstructed 3-D hand geometry and classified
into discrete hand states. The signing vocabulary contains
\emph{open}, \emph{fist}, \emph{one}, \emph{two}, \emph{three},
\emph{four}, and \emph{thumb}, while reach-to-grasp motion uses an
additional task-specific \emph{grasp} configuration.

For signing sequences, a Viterbi-style temporal decoder improves the
consistency of handshape transitions~\cite{viterbi1967}. For
reach-to-grasp sequences, the detected hand state determines the
corresponding grasp preset. Each state is mapped to six calibrated
actuation channels, which are subsequently converted through the
robot hand kinematics.

After frame-wise arm fitting and hand-state assignment, a symmetric
Gaussian filter conditions the arm and hand commands. The current
implementation uses $\sigma=1.25$ frames. For joint sequence
$\mathbf q_t$,
\begin{equation}
\tilde{\mathbf q}_t=\sum_{k=-K}^{K}g_k\mathbf q_{t-k},\qquad
g_k=\frac{\exp(-k^2/2\sigma^2)}
{\sum_{j=-K}^{K}\exp(-j^2/2\sigma^2)}.
\label{eq:gaussian}
\end{equation}
The symmetric kernel avoids a one-sided delay and mainly suppresses
high-frequency variation from frame-wise numerical optimization. The
time axis is then linearly resampled to the desired execution and
visualization speed, producing the final robot command trajectory.

\section{Experiments}
\label{sec:experiments}

The evaluation comprises 16 fixed-camera monocular videos, including
10 signing and six reach-to-grasp sequences, with 1,769 source frames
in total. All sequences are processed through the same reconstruction
and retargeting pipeline. Metrics are first computed within each
sequence and subsequently aggregated across sequences. Confidence
intervals are obtained through sequence-level bootstrap resampling,
treating each video as the unit of statistical evidence.

\subsection{Experimental Setup}

Experiments use a fixed-base Haier S01 equipped with two 7-DoF arms
and two six-channel dexterous hands, yielding 26 command values per
frame.
Reach-to-grasp objects are placed at calibrated corresponding
locations in the demonstration and robot workspaces. Five signing
sequences contain evaluable bilateral wrist relations.

Human reconstruction is compared with frame-wise SAM 3D Body,
SAM-Body4D, and late Dyn-HaMR hand attachment. Retargeting
comparisons include absolute-position transfer, independent-arm
geometry, bilateral wrist constraints, relative palm calibration,
and Gaussian command conditioning. The common output stage for
retargeting comparisons is the Gaussian-conditioned trajectory.
Isaac validation uses kinematic joint-state playback.

\subsection{Reconstruction Results}

\begin{table*}[t]
\centering
\caption{Human reconstruction results over 16 sequences and 1,769
source frames. Hand 2-D error is in pixels, local 3-D residual is
dimensionless, and second-order displacement is in mm/frame$^2$.}
\label{tab:reconstruction}
\footnotesize
\begin{tabular}{lcccc}
\toprule
Method & Hand 2-D (px) & Local 3-D & 2nd-order disp. & Round-trip outliers\\
 & Mean / P95 & Mean / P95 & Mean / P95 / max & Count\\
\midrule
SAM 3D Body & 22.36 / 68.84 & 0.270 / 0.607 &
4.85 / 15.14 / 174.64 & 75\\
SAM-Body4D & 23.84 / 71.08 & 0.284 / 0.653 &
5.02 / 16.92 / 101.45 & 94\\
Late fusion & 21.07 / 50.89 & N/A$^\dagger$ &
5.90 / 21.88 / 260.83 & 302\\
\textbf{Unified MHR} & \textbf{7.21 / 20.48} &
\textbf{0.090 / 0.213} & \textbf{3.34 / 13.80 / 164.17} &
\textbf{12}\\
\bottomrule
\multicolumn{5}{l}{\scriptsize
$^\dagger$The local 3-D residual is not defined for late fusion because
Dyn-HaMR local hand geometry is directly grafted onto the body.}
\end{tabular}
\end{table*}

Across the 16 sequences, the final unified MHR reconstruction
achieved a mean hand reprojection error of 7.21~px, compared with
22.36~px for frame-wise SAM 3D Body and 23.84~px for SAM-Body4D
(Table~\ref{tab:reconstruction}). The corresponding P95 error was
20.48~px, down from 68.84~px for SAM 3D Body. The mean reduction is
approximately 67.8\%. The final normalized local 3-D residual was
0.090, compared with 0.270 for SAM 3D Body and 0.284 for SAM-Body4D,
an improvement of approximately 66.7\% over SAM 3D Body. A
20,000-sample paired bootstrap using video sequences as
the resampling unit estimated a 17.95~px reduction in sequence-level
mean hand reprojection error relative to SAM 3D Body, with a 95\%
confidence interval of 12.61--23.25~px. The paired improvement in
normalized local 3-D residual was 0.187 (95\% confidence interval:
0.162--0.215). The late-fusion baseline directly attaches Dyn-HaMR
local hand geometry to the reconstructed body; its local 3-D
residual is therefore omitted from the comparison.

Before model-space repair, unified MHR yielded mean and P95 local 3-D
residuals of 0.088 and 0.209. The mean, P95, and maximum second-order
displacements were 3.54, 14.46, and 205.51~mm/frame$^2$, and 138
round-trip outliers were detected. After repair, the local 3-D
residuals were 0.090 and 0.213. The second-order displacements were
3.34, 13.80, and 164.17~mm/frame$^2$, and the outlier count was 12.
This represents a 91.3\% reduction in detected round-trip failures
with little change in local 3-D agreement.

\subsection{Ablation Study}

\begin{table*}[t]
\centering
\caption{Ablation results with sequence-level paired bootstrap
confidence intervals. Positive values denote improvements after
introducing the corresponding component.}
\label{tab:ablation}
\footnotesize
\begin{tabular}{llcc}
\toprule
Component & Metric & $N$ & Paired improvement [95\% CI]\\
\midrule
Relative geometry & Forearm error ($^\circ$) & 16 &
5.57 [4.21, 7.13]\\
Relative geometry & Palm error ($^\circ$) & 16 &
3.48 [0.23, 7.02]\\
Palm calibration & Palm error ($^\circ$) & 16 &
14.04 [9.82, 18.57]\\
Bilateral relation & Wrist separation & 5 &
0.004 [0.002, 0.006]\\
Bilateral relation & Height-order agreement & 5 &
0.313 [0.115, 0.512]\\
\bottomrule
\end{tabular}
\end{table*}

Across five bilateral signing sequences, independent-arm fitting
preserved wrist-height ordering in 221 of 318 evaluable frames.
Adding the bilateral relation increased agreement to 318/318 and
reduced normalized wrist-separation error from 0.157 to 0.153
(Table~\ref{tab:ablation}). The paired sequence-level gain in
height-order accuracy was 0.313 (95\% confidence interval:
0.115--0.512).

Relative palm calibration reduced paired palm-orientation error by
14.04$^\circ$ (95\% confidence interval:
9.82--18.57$^\circ$). Compared with absolute-position
retargeting, the complete geometric formulation reduced
sequence-level mean forearm error by 5.57$^\circ$ (95\% confidence
interval: 4.21--7.13$^\circ$) and palm error by 3.48$^\circ$
(95\% confidence interval: 0.23--7.02$^\circ$).
The corresponding mean forearm error decreased from 12.24$^\circ$ to
6.67$^\circ$, while mean palm error decreased from 22.28$^\circ$ to
18.79$^\circ$.

\subsection{Temporal Analysis}

\begin{table}[!htbp]
\centering
\caption{Gaussian command conditioning over all 16 sequences.
Values aggregate one statistic per sequence. Joint increments are
in degrees; second-order differences are in degrees/frame$^2$.}
\label{tab:temporal}
\scriptsize
\begin{tabular}{lccc}
\toprule
Metric & Before & Gaussian & Paired reduction [95\% CI]\\
\midrule
Max joint step & 17.09 & 10.49 & 6.60 [4.56, 8.84]\\
P95 2nd difference & 4.03 & 1.40 & 2.63 [2.15, 3.15]\\
Max 2nd difference & 18.61 & 3.67 & 14.94 [11.46, 18.96]\\
\bottomrule
\end{tabular}
\end{table}

Gaussian conditioning reduced the sequence-level maximum joint
increment by 6.60$^\circ$ (95\% confidence interval:
4.56--8.84$^\circ$), the P95 second-order difference by
2.63$^\circ$/frame$^2$ (95\% confidence interval: 2.15--3.15),
and the maximum second-order difference by
14.94$^\circ$/frame$^2$ (95\% confidence interval: 11.46--18.96;
Table~\ref{tab:temporal}). The improvement in temporal regularity
introduced only minor changes in geometric fidelity:
$-0.03^\circ$, $+0.10^\circ$, and $+0.33^\circ$ for the upper
arm, forearm, and palm, respectively.

\subsection{Task-Level Evaluation}

\begin{figure*}[t]
\centering
\includegraphics[width=0.784\textwidth]{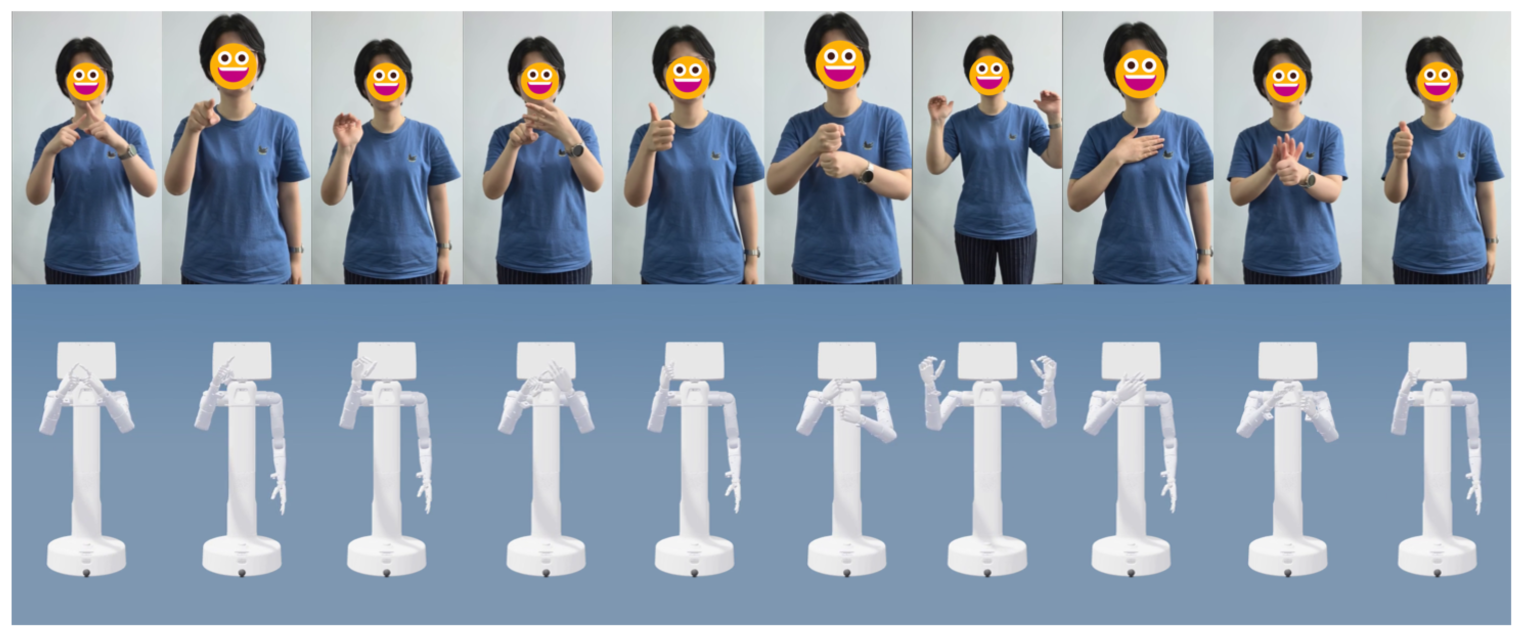}
\caption{Retargeting from monocular human video to robot kinematic
simulation for ten representative signs: \emph{person}, \emph{you},
\emph{can}, \emph{name}, \emph{good}, \emph{work}, \emph{meeting},
\emph{i}, \emph{love}, and \emph{thank you}. Each column shows a
source human pose above its retargeted robot pose.}
\label{fig:retargeting_results}
\end{figure*}

Figure~\ref{fig:retargeting_results} compares corresponding human and
simulated robot poses for ten signs. The examples cover unilateral and
bilateral motion, varied elbow flexion, changing wrist relations, palm
orientations, and hand configurations. The robot reconstructs arm
directions, elbow configuration, bilateral wrist relations, and
relative palm orientation using its own link lengths and kinematic
structure.

\textbf{Sign-language motion transfer:}
The 10 signing sequences are evaluated through arm, elbow, and palm
geometry and kinematic playback. In the five bilateral sequences,
the bilateral constraints preserved the source wrist-height
relation in all 318 evaluable frames. Discrete robot hand
configurations accompany the transferred arm and palm motion.

\textbf{Reach-to-grasp motion transfer:}
Across the six reach-to-grasp sequences, the mean upper-arm, forearm,
elbow-bending, elbow-flexion, and palm-orientation errors were
6.02$^\circ$, 8.07$^\circ$, 6.57$^\circ$, 3.68$^\circ$, and
24.10$^\circ$, respectively. Mean wrist approach-direction error was
16.66$^\circ$, and the application-specific palm-orientation error
was 23.95$^\circ$. Wrist-displacement error averaged 129.68~mm at
the detected onset of the grasp state and 52.78~mm at the final
frame. Relative-geometric transfer reduced palm-orientation error
by 11.23$^\circ$ relative to absolute-position retargeting while
increasing final wrist-displacement error by 12.99~mm.

\subsection{Simulation Validation}

All 16 retargeted trajectories completed kinematic playback in
Isaac with finite joint states and zero model joint-limit
violations. Offline geometric validation showed valid cross-arm
clearance in 15 of the 16 sequences.

\section{Real-Robot Demonstrations}
\label{sec:real_robot}

We deployed the complete reconstruction and retargeting pipeline on a
Haier S01 dual-arm robot. Three representative motions were selected:
reach-to-grasp, the sign \emph{person}, and the sign \emph{thank you}.
They cover object-directed unilateral motion, coordinated bilateral
configuration, and signing motions with distinct hand postures.

\begin{figure}[t]
\centering
\includegraphics[width=\columnwidth]{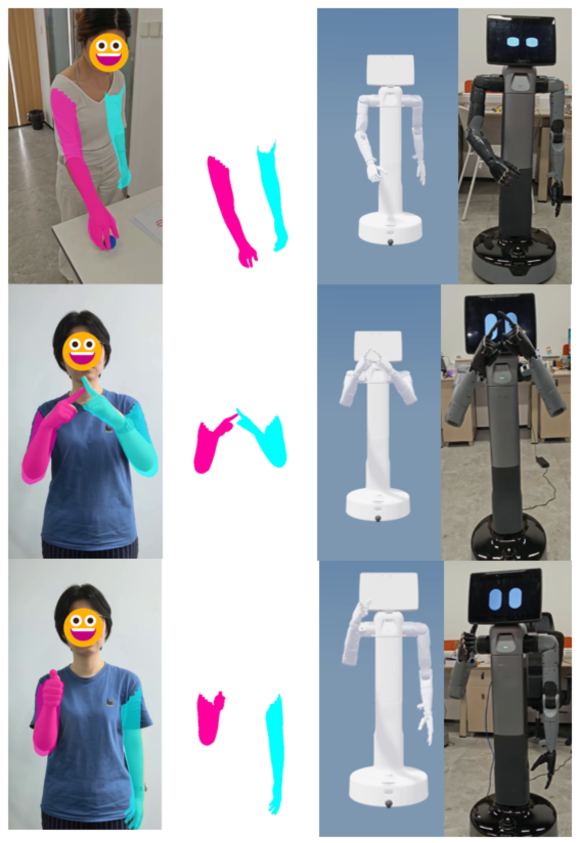}
\caption{Representative validation from monocular human video to a
physical robot. The rows show reach-to-grasp, \emph{person}, and
\emph{thank you}. From left to right, each row presents a source video
frame, body--hand 3-D reconstruction, robot kinematic simulation, and
physical execution.}
\label{fig:figure_4}
\end{figure}

Figure~\ref{fig:figure_4} shows the complete correspondence from human
demonstration to physical execution. Each motion passes through
monocular-video input, unified body--hand reconstruction,
cross-embodiment geometric retargeting, and robot command generation.
The simulated and physical robots use the same joint-command
representation. This shared representation enables direct qualitative
comparison between simulated and physical execution.

The reach-to-grasp sequence evaluates physical realization of
unilateral motion. The shoulder drives the upper arm and forearm toward
the target while the wrist and hand change posture. Retargeting uses
the human upper-arm direction, forearm direction, elbow configuration,
and palm orientation to construct the robot arm pose. A calibrated
grasp handshape adapts distal motion. The physical execution preserves
the principal arm-extension direction and wrist posture observed in
simulation, consistent with the intended cross-embodiment geometric
mapping.

The sign \emph{person} further tests bilateral geometry. It involves
flexion of both arms, relative wrist placement, and two local
handshapes. Each arm first realizes its local direction and elbow
configuration. Bilateral constraints then coordinate the relative
wrist positions, after which palm orientation and discrete handshape
complete the pose. The observed correspondence between simulation and
physical execution is consistent with the bilateral geometric
constraints imposed during retargeting.

The sign \emph{thank you} is dominated by one upper limb and includes
clear changes in elbow flexion, forearm direction, and hand
configuration. Unlike reaching, its goal is to preserve expressive
posture rather than attain an object position. The physical result
shows that the same geometric representation transfers this different
motion type without changing the reconstruction or robot-kinematics
pipeline.

Together, the three demonstrations cover object-directed unilateral
motion, bilateral coordination, and posture-oriented signing. The
physical robot reproduces the main arm geometry and hand states formed
in simulation. These results qualitatively validate the complete path
from ordinary monocular video through unified body--hand reconstruction
and cross-embodiment retargeting to physical execution.

\section{Conclusion}
\label{sec:conclusion}

This article presented a geometry-preserving framework for
transferring upper-body and hand motion from monocular human video
to a morphologically different dual-arm robot. Unified body--hand
reconstruction jointly constrains frame-wise body estimates,
video-level observations, and detailed hand observations in one
differentiable MHR state. Model-space repair improves the temporal
stability of distal motion. Relative arm geometry, calibrated palm
orientation, and bilateral wrist relations then define
morphology-independent targets for multi-stage inverse kinematics.
Robot-specific handshapes and temporal conditioning complete the
dexterous-hand adaptation and command sequence.

On 16 monocular sequences containing 1,769 source frames, unified MHR
reconstruction reduced mean hand reprojection error from 22.36 to
7.21~px. Model-space repair reduced detected round-trip hand outliers
from 138 to 12. Relative geometry reduced sequence-level mean forearm
and palm errors by 5.57$^\circ$ and 3.48$^\circ$ compared with
absolute-position transfer. Bilateral wrist constraints improved the
preservation of relative wrist configuration. Gaussian conditioning
reduced the maximum adjacent-frame joint increment from 17.09$^\circ$
to 10.49$^\circ$ while introducing only small geometric changes. All
16 trajectories completed Isaac kinematic playback with finite joint
states and no model joint-limit violations.

Signing and reach-to-grasp evaluate complementary forms of upper-body
motion. The former emphasizes bilateral coordination, palm
orientation, and handshape, whereas the latter emphasizes coordinated
arm approach, wrist orientation, and grasp configuration. Physical
experiments with reach-to-grasp, \emph{person}, and \emph{thank you}
show that the same command interface transfers motions from simulation
to the Haier S01 robot. The resulting system forms a complete pipeline
from ordinary monocular video to unified body--hand reconstruction,
cross-embodiment geometric retargeting, and physical robot execution.
Future work will address moving cameras, continuous dexterous-finger
motion, explicit object and contact constraints, and additional robot
morphologies.

\bibliographystyle{IEEEtran}
\bibliography{references}

\end{document}